\documentclass[sigconf]{acmart}
\usepackage{xcolor,colortbl}

\usepackage{graphicx}
\usepackage{balance}  
\usepackage{graphics}
\usepackage{mdwlist}
\usepackage{xspace}
\usepackage[ruled, vlined, linesnumbered]{algorithm2e} 
\usepackage{booktabs}
\usepackage{soul}

\usepackage{amsmath}
\usepackage{amsfonts}
\usepackage{amsbsy}
\usepackage{amsthm}
\usepackage{caption}
\usepackage{subcaption}
\usepackage{multirow}
\usepackage{mathrsfs}
\usepackage[mathscr]{eucal} 
\usepackage{paralist} 
\usepackage{mathtools}
\usepackage{verbatim}
\usepackage{tabularx}
\usepackage{makecell}
\usepackage{hhline}
\usepackage{placeins}
\usepackage{diagbox}
\usepackage{enumitem}
\usepackage{pifont}
\theoremstyle{definition}

\theoremstyle{remark}

\let\oldnl\nl
\newcommand{\nonl}{\renewcommand{\nl}{\let\nl\oldnl}}
\usepackage{dsfont}

\usepackage{array}

\usepackage{enumitem}
\AtBeginDocument{%
  \providecommand\BibTeX{{%
    \normalfont B\kern-0.5em{\scshape i\kern-0.25em b}\kern-0.8em\TeX}}}

\setcopyright{acmcopyright}
\copyrightyear{2022}
\acmYear{2022}

\acmConference[SIGIR '22] {Proceedings of the 45th International ACM SIGIR Conference on Research and Development in Information Retrieval}{July 11--15, 2022}{Madrid, Spain.}
\acmBooktitle {Proceedings of the 45th International ACM SIGIR Conference on Research and Development in Information Retrieval (SIGIR '22), July 11--15, 2022, Madrid, Spain}
\acmPrice{15.00}
\acmISBN{978-1-4503-8732-3/22/07}
\acmDOI{10.1145/XXXXXX.XXXXXX}

\newcommand\best[1]{\textbf{#1}}

\definecolor{myyellow}{RGB}{255,255,100}
\definecolor{mygray}{RGB}{220,220,220}

\definecolor{darkgreen}{RGB}{0,128,0}
\definecolor{mypurple}{RGB}{148,0,211}

\newcolumntype{a}{>{\columncolor{mygray}}c}
\newcolumntype{g}{>{\columncolor{mygray}}c}

\newlist{Properties}{enumerate}{2}
\setlist[Properties]{label=Property \arabic*., font=\textbf, itemindent=*}
\newcommand{\smallsection}[1]{{\noindent {\bf{\underline{\smash{#1}}}}}}

\newtheorem{problem}{\textbf{Problem}}

\definecolor{myred}{RGB}{195, 79, 82}
\definecolor{mygreen}{RGB}{10, 130 10}  
\definecolor{myblue}{RGB}{74, 113 175}
\begin{document}
\fancyhead{}

	
    \title{AHP: Learning to Negative Sample for Hyperedge Prediction}
    \settopmatter{authorsperrow=4}
    \author{Hyunjin Hwang}
	\authornote{Equal Contribution.}
	\affiliation{%
		\institution{KAIST AI}
		\city{Seoul}
		\country{South Korea}
	}
	\email{hyunjinhwang@kaist.ac.kr}
	
	\author{Seungwoo Lee}
	\authornotemark[1]
	\affiliation{%
		\institution{KAIST EE}
		\city{Daejeon}
		\country{South Korea}
	}
	\email{ksalsw1996@kaist.ac.kr}
	
	\author{Chanyoung Park}
	\affiliation{%
		\institution{KAIST ISysE \& AI}
		\city{Daejeon}
		\country{South Korea}
	}
	\email{cy.park@kaist.ac.kr}
	
	\author{Kijung Shin}
	\affiliation{%
		\institution{KAIST AI \& EE}
		\city{Seoul}
		\country{South Korea}
	}
	\email{kijungs@kaist.ac.kr}
    
    \begin{abstract}
    Hypergraphs (i.e., sets of hyperedges) naturally represent group relations 
(e.g., researchers co-authoring a paper and ingredients used together in a recipe), each of which corresponds to a hyperedge (i.e., a subset of nodes).
Predicting future or missing hyperedges bears significant implications for many applications (e.g., collaboration and recipe recommendation).
What makes hyperedge prediction particularly challenging is the vast number of non-hyperedge subsets, which grows exponentially with the number of nodes.
Since it is prohibitive to use all of them as negative examples for model training, it is inevitable to sample a very small portion of them, and to this end, heuristic sampling schemes have been employed.
However, trained models suffer from poor generalization capability for examples of different natures.
In this paper, we propose AHP, an adversarial training-based hyperedge-prediction method. It learns to sample negative examples without relying on any heuristic schemes. 
Using six real hypergraphs, we show that AHP generalizes better to negative examples of various natures. It yields up to \textbf{28.2\% higher AUROC} than the best existing methods and often even outperforms its variants with sampling schemes tailored to test sets.
    \end{abstract}


    
    \maketitle
    
    \section{Introduction}
    \label{sec:intro}
    \begin{figure}[t]
    \centering
    \includegraphics[width=0.99\linewidth]{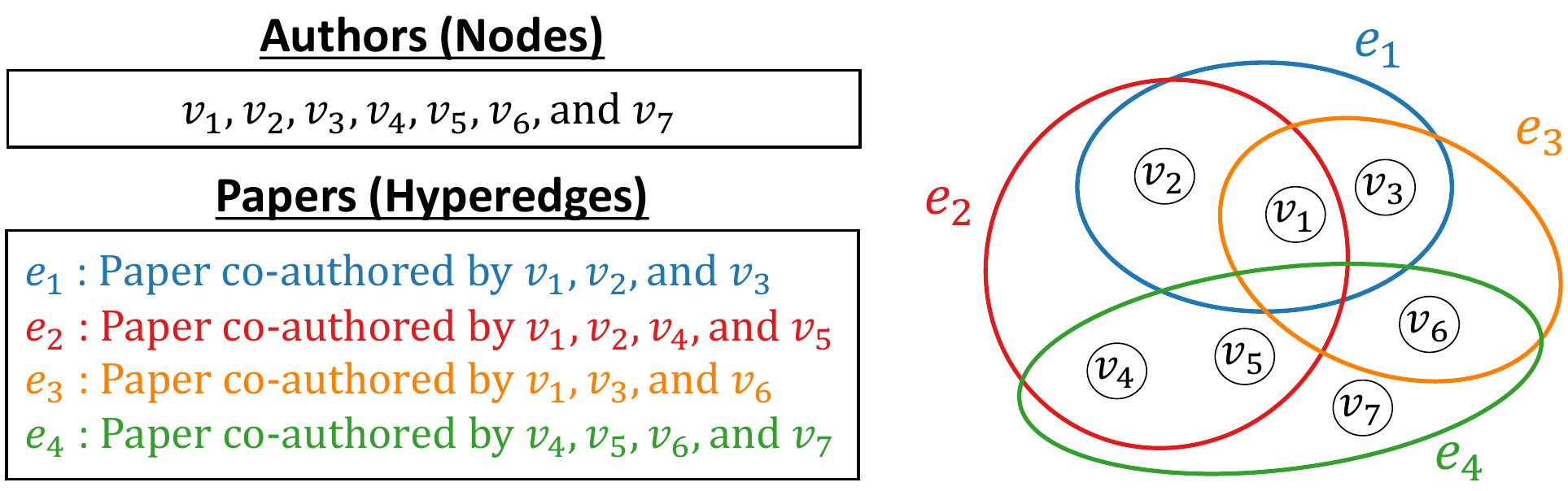} \\
        \vspace{-0.8mm}
    \caption{Collaboration data modeled as a hypergraph.}
    \label{fig:example}
\end{figure}

\begin{figure}[t]
    \vspace{-4.9mm}
    \centering
      \includegraphics[width=0.9\linewidth]{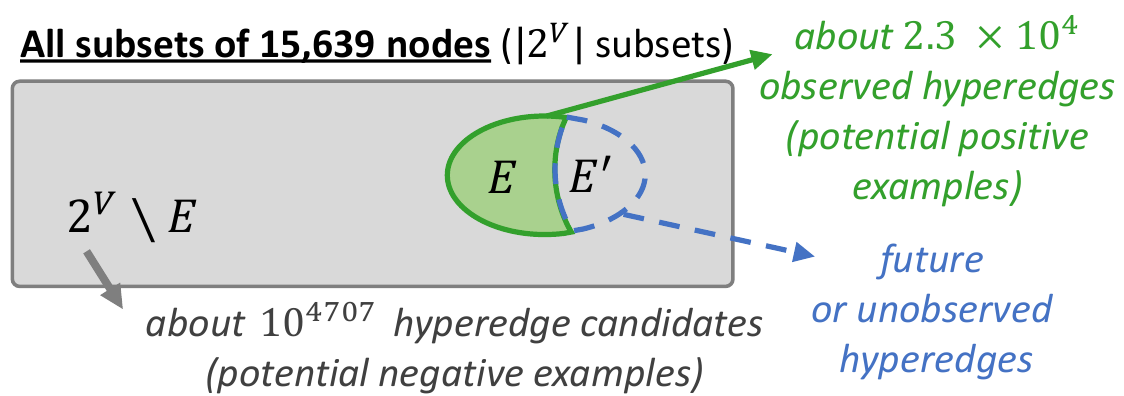} \\
        \vspace{-0.8mm}
    \caption{Venn diagram of hyperedges and candidates in the \textsf{DBLP} dataset. The number of candidates is extremely large.}\label{fig:Venn}
\end{figure}

Group relations among objects are common in complex systems: (a) researchers co-authoring a paper, (b) items co-purchased by a shopper, (c) ingredients in a recipe, (d) proteins inducing a chemical reaction together, (e) drugs causing side effects when taken together, to name a few \cite{amburg2020clustering,benson2018sequences,benson2018simplicial,comrie2021hypergraph,choo2022persistence,do2020structural,
choe2022midas,lee2021thyme,kook2020evolution,lee2021hyperedges,lee2020hypergraph,torres2021and}. They together are naturally modeled as a \textit{hypergraph} (i.e., a set of hyperedges) where each \textit{hyperedge} (i.e., a subset of nodes) indicates a group relation among the nodes in it. See Fig.~\ref{fig:example} for an example hypergraph. 

\textit{Hyperedge prediction} (i.e., the task of predicting future or unobserved hyperedges) is a fundamental task with numerous applications, including collaboration/movie/recipe recommendation \cite{liu2018context,yu2018modeling,zhang2018beyond}, chemical reaction prediction \cite{zhang2018beyond, nhp}, drug side-effect prediction \cite{vaida2019hypergraph,nguyen2021centsmoothie}.
Thus, it has received considerable attention in the recommendation system and machine learning communities.

One of the main challenges in hyperedge prediction is the vast number of potential negative examples, i.e., subsets of nodes that are not a hyperedge.
As illustrated in Fig.~\ref{fig:Venn}, their number is about $10^{4,707}$ in a dataset used, and formally, their number is $2^{|V|}-|E|$, where $V$ and $E$ are the sets of nodes and hyperedges, respectively. 
Thus, it is practically impossible to utilize all non-hyperedge subsets as negative examples when training machine learning models. 

Therefore, it is inevitable to sample a tiny portion of potential negative examples, and to this end, existing machine-learning-based hyperedge-prediction methods \cite{tu2018structural,zhang2019hyper,yoon2020much,nhp,patil2020negative} resort to heuristic sampling schemes. 
For example, a negative example is sampled by replacing half of the nodes in a hyperedge with random nodes \cite{nhp}.

However, we observe that models trained using such negative examples suffer from poor generalization capability.
Specifically, their capability of classifying positive and negative examples heavily depends on sampling schemes used for training and test sets. 
Our finding obtained from state-of-the-art methods \cite{nhp,yoon2020much,zhang2019hyper}
complements the findings in \cite{patil2020negative}, where rule-based and simple learning-based techniques are used to
compare sampling schemes.

In this paper, we propose AHP (\textbf{A}dversarial training-based \textbf{H}yper-edge \textbf{P}rediction), which learns to sample negative examples without relying on heuristic sampling schemes. Its generator aims to fool its discriminator by sampling hard negative examples.
We adapt a hypergraph neural network for the discriminator and design a generator simple yet well-suited to hyperedge prediction, while our training method is model agnostic. 
Based on experimental results, we summarize our contributions as follows:

\vspace{-0.5mm}
\begin{itemize}[leftmargin=*]
    \item {\bf Observation:} We show that heuristic sampling schemes limit the generalization ability of deep learning-based hyperedge-prediction.
    \item {\bf Solution:} 
    AHP learns to sample negative examples by adversarial training for better generalization.
    In terms of AUROC, AHP is up to \textbf{28.2\% better} than best existing methods and up to \textbf{5.5\% better} than  variants with sampling schemes tailored to test sets.
    \item {\bf Experiments:} 
    We compare AHP with 3 sampling schemes and 3 recent hyperedge-prediction methods on 6 real hypergraphs.
\end{itemize}
\vspace{-0.5mm}

\noindent For \textbf{reproducibility}, the code and datasets used in the paper are available at \url{https://github.com/HyunjinHwn/SIGIR22-AHP}.

    \section{Preliminaries}
    \label{sec:prelim}
    In this section, we provide some preliminaries on hyperedge prediction and hypergraph neural networks.

\smallsection{Basic concepts.}
Consider a \textbf{hypergraph} $H=(V,E)$ consisting of a set of nodes $V=\{v_1, \cdots, v_{|V|}\}$ and a set of hyperedges $E=\{e_1, \cdots, e_{|E|}\}$ (see Fig.~\ref{fig:example} for an example).
Each \textbf{hyperedge} $e_i\in E$ is a subset of nodes, i.e., $e_i\subseteq V$. 
Each $(i,j)$-th entry of the \textbf{incidence matrix} $A\in \mathbb{R}^{|V| \times |E|}$ of $H$ is $1$ if $v_i\in e_j$, and it is $0$ otherwise.
We also assume \textbf{node features} $X\in \mathbb{R}^{|V| \times d}$, where each $i$-th row corresponds to the $d$-dimensional feature of the node $v_i$.

\smallsection{Problem definition.}
For a given hypergraph $H=(V,E)$,
the objective of hyperedge prediction is to find the \textbf{target set} $E'\subseteq 2^V\setminus E$, which typically consists of (a) unobserved hyperedges or (b) new hyperedges that will arrive in the near future. Each member of $2^V\setminus E$ is called a \textbf{hyperedge candidate} as it may belong to $E'$.
Due to the vast number of hyperedge candidates (see Fig.~\ref{fig:Venn} for an example), ranking or evaluating every candidate is computationally infeasible. Thus, hyperedge prediction is typically formulated as a classification problem below \cite{tu2018structural,zhang2019hyper,yoon2020much,nhp,patil2020negative}.

\begin{problem}[Hyperedge prediction]\label{def:HEpred}
    Given a hypergraph $H=(V,E)$, node features $X\in \mathbb{R}^{|V| \times d}$, and a hyperedge candidate $c\in2^V\setminus E$, we aim to classify whether $c$ belongs to the target set $E'$ or not.
\end{problem}

\begin{figure}[t]
    \vspace{-1mm}
    \centering
    \includegraphics[width=\linewidth]{{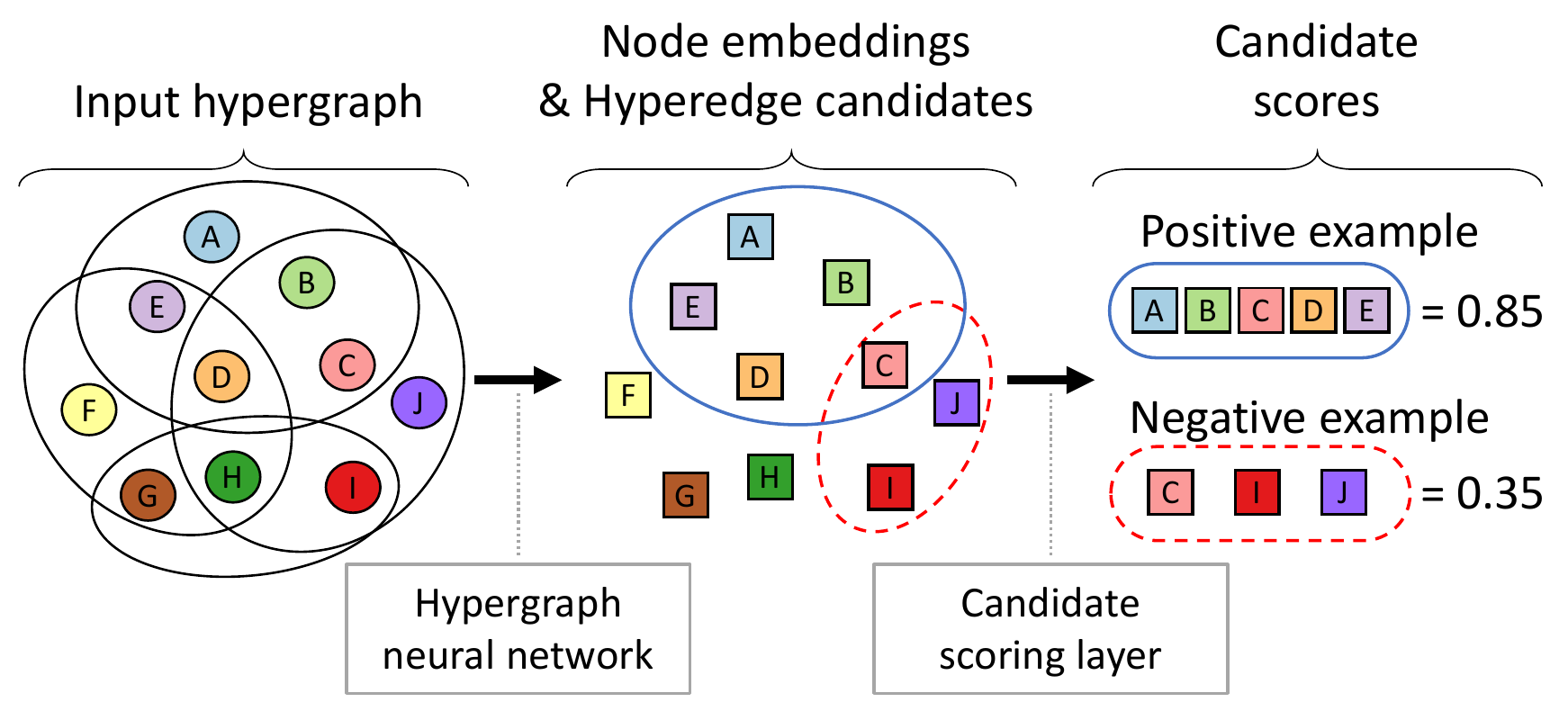}}
    \caption{Hyperedge prediction using a hypergraph neural network and a candidate scoring layer.}
    \label{fig:hepred}
\end{figure}

\smallsection{Hypergraph neural networks.}
Hypergraph neural networks (HyperGNNs) \cite{tu2018structural,feng2019hypergraph,hypergcn,dong2020hnhn,nhp} are a class of neural networks designed to perform inference on hypergraphs, and they proved effective for a number of applications \cite{nhp, li2021hyperbolic, wang2020next, he2021click, zhang2021double, JointRecommendation}.
A representative HyperGNN is HNHN~\cite{dong2020hnhn}, which repeats (a) producing node embeddings by aggregating the embeddings of incident hyperedges and (b) producing hyperedge embeddings by aggregating the embeddings of incident nodes by using Eq.~\eqref{eq:hnhn} for each layer $l\in \{1,\cdots, L\}$.
\begin{equation}
    \small
    X^{(l)}_{E}=\sigma (A^TX^{(l-1)}_VW^{(l)}_V+b^{(l)}_V), \ \ \ X^{(l)}_V=\sigma (AX^{(l)}_EW^{(l)}_E+b^{(l)}_E), \label{eq:hnhn}
\end{equation}
where {\small $W^{(l)}_V$} and {\small $W^{(l)}_E$} are learnable weight matrices and {\small $b^{(l)}_V$} and {\small $b^{(l)}_E$} are learnable bias matrices.
The matrix $A$ is the incident matrix, and the initial embeddings {\small $X^{(0)}_V$} equal to the given node features $X$. The function $\sigma$ is a nonlinear activation function, and in Eq.~\eqref{eq:hnhn}, normalization terms are omitted for simplicity.
The output of HNHN is {\small $X^{(L)}_V$}, i.e., the node embeddings in the last layer, as in Fig.~\ref{fig:hepred}.

\smallsection{Candidate scoring.}
In order to apply hypergraph neural networks, whose output is node embeddings, to hyperedge prediction, two additional steps are required.
For a given hyperedge candidate $c\in 2^V\setminus E$, the embeddings of nodes in $c$ are pooled (for example, by being averaged) into the embedding of $c$. Then, the embedding is used (for example by being fed into MLP) to estimate how likely $c$ belongs to the target set $E'$, as in Fig.~\ref{fig:hepred}.

\smallsection{Training and negative sampling.} 
Based on the assumption that (unobserved or future) hyperedges in the target set $E'$ bear similarities with (observed) hyperedges in $E$, the members of $E$ are used as \textbf{positive examples}, and the members of $2^V\setminus E$ are used as \textbf{negative examples} during training.
That is, the hypergraph neural network and the hyperedge scoring layer are trained so that positive examples earn higher scores and negative examples earn lower scores, as in Fig.~\ref{fig:hepred}.
Due to the vast size of $2^V\setminus E$ (see Fig.~\ref{fig:Venn} for an example), it is prohibitive to use all its members as negative examples, and it is inevitable to sample a tiny portion of them. This sampling process is called \textbf{negative sampling}. As discussed in Sect.~\ref{sec:intro}, to this end, heuristic schemes are commonly used, such as:
\begin{itemize}[leftmargin=*]
    \item \textbf{Sized negative sampling (SNS)}: fill a set with $k$ random nodes.
    \item \textbf{Motif negative sampling (MNS)}: fill a set of size $k$ by repeatedly merging it with a random incident edge in the clique expansion.
    \item \textbf{Clique negative sampling (CNS)}: pick a random hyperedge and replace a random constituent node with a random node that is adjacent to all other constituent nodes.
\end{itemize}
In all the sampling schemes above \cite{patil2020negative}, random selection is uniform, and the size $k$ of each negative example is drawn iid from the hyperedge size distribution in the input hypergraph.
As discussed in Sect.~\ref{sec:intro} and demonstrated in Sect.~\ref{sec:exp}, these sampling schemes limit
the generalization capability of hyperedge-prediction methods.
    
    \section{Proposed Method}
    \label{sec:method}
\begin{figure}[t]
    \centering
    \includegraphics[width=0.99\linewidth]{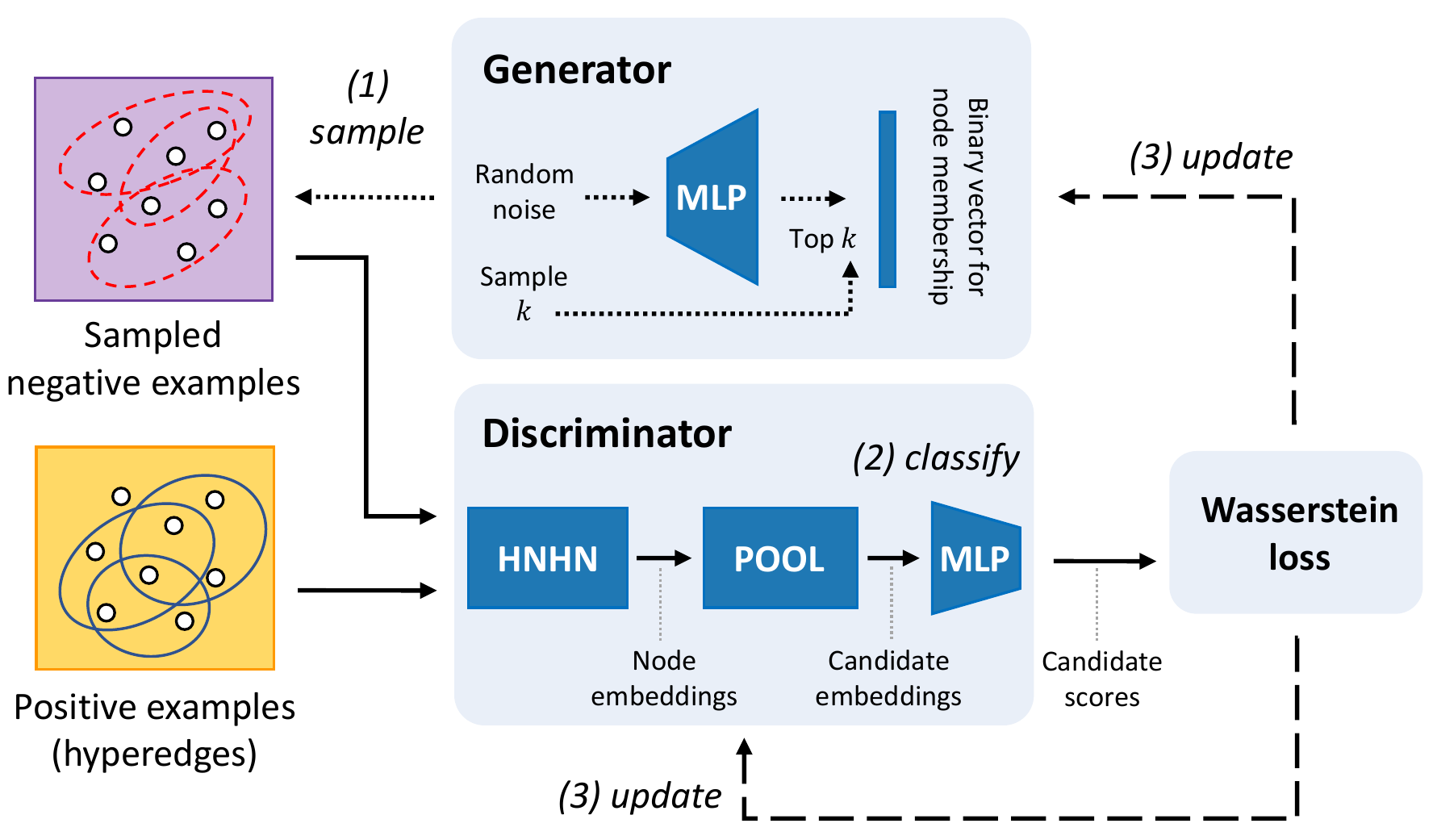}
    \caption{\label{fig:ahp} Training of AHP. The generator learns to sample negative examples, aiming at fooling the discriminator.
    }
\end{figure}

    In this section, we propose AHP (\textbf{A}dversarial training-based \textbf{H}yper-edge \textbf{P}rediction), which learns to negative sample, without relying on heuristic negative sampling schemes, for deep learning-based hyperedge prediction.
AHP consists of (a) a \textbf{generator} for sampling negative examples and (b) a \textbf{discriminator} for candidate scoring, which are \textbf{trained adversarially}, as depicted in Fig.~\ref{fig:ahp}.

\smallsection{Challenges.}
It should be noted that our objective of adversarial training is different from the typical one (i.e., to train a generator).  
For successful hyperedge prediction, the power of the generator needs to be within an appropriate range. If the generator is too weak, it produces only trivial negative samples, which are not helpful to effective training of the discriminator.
On the other hand, 
if it  is powerful enough to produce positive examples or those in the target set, it loses its role of sampling negative examples. 

\smallsection{Generator.}
Due to this reason, for the generator of AHP, using our simple architecture (described below) resulted in more accurate hyperedge prediction than adapting advanced architectures (e.g., \cite{vinyals2015pointer}), in our preliminary study.
The generator of AHP consists of a simple multi-layer perceptron (MLP) with three layers and LeakyReLU as activation functions between them.
Specifically, the generator first draws $k$, the size of a negative example, from the size distribution of positive examples.
Then, its MLP model generates a vector for node membership from a random Gaussian noise. 
Lastly, it selects $Top$-$k$ nodes in the vector that together compose a negative example, which is then fed into the discriminator.

\smallsection{Discriminator.}
Given a subset of nodes (i.e., a positive or negative example during training and a hyperedge candidate during inference), the discriminator measures a score that indicates how likely the nodes together form a hyperedge.
For the discriminator, we use HNHN for node embedding and the $maxmin$ function (i.e., element-wise maximum - element-wise minimum) followed by MLP for candidate scoring (see Sect.~\ref{sec:prelim} for HNHN and candidate scoring).
Intuitively, the $maxmin$ function measures how diverse the embeddings of nodes are in a given subset, and the diversity may reveal important clues regarding hyperedge formation as homophily (i.e., similar nodes are more likely to form hyperedges together than dissimilar nodes) is pervasive in real-world hypergraphs \cite{lee2021hyperedges}. 

\smallsection{Adversarial training.}
As depicted in Fig.~\ref{fig:ahp}, we train the generator $G$ and the discriminator $D$ adversarially, by repeating three steps for each batch $S$ of positive examples,
(1) \textbf{sample} $|S|$ negative examples using $G$,
(2) \textbf{classify} the positive and negative examples using $D$, and 
(3) \textbf{update} $G$ and $D$ using gradient descent on their losses.
As $D$ aims to make the scores of positive examples larger and those of negative examples smaller, for a given hypergraph $H$ and node features $X$, the loss function for $D$ is:
\begin{equation}
    \small
    \mathcal{L}_D = -\frac{1}{|S|}\sum\nolimits_{s\in S}[D(s|H, X)] + \frac{1}{|S|}\sum\nolimits_{j=1}^{|S|}[D(G(z_j)|H, X))], \label{eq:dloss}
\end{equation}
where $G(z_j)$ is a negative example generated from the noise $z_j$.
Since $G$ aims to fool $D$ and make it classify its negative examples as positive,
the loss function for $G$ is:
\begin{equation}
    \small
    \mathcal{L}_G = -\frac{1}{|S|}\sum\nolimits_{j=1}^{|S|}[D(G(z_j)|H, X))]. \label{eq:gloss}
\end{equation}

\smallsection{Memory bank for training stability.}
Adversarial training often suffers from instability \cite{arjovsky2017wasserstein,arjovsky2017towards,roth2017stabilizing,thanh2019improving}, and to mitigate it, we employ a memory bank. Specifically, we maintain the ``hardest'' negative examples, i.e. negative examples that earned the highest scores by the discriminator in the previous iteration, and reuse them as negative examples.\footnote{With the extra negative samples $M$, the loss for $D$ is
$\mathcal{L}_D =  -\frac{1}{|S|}\sum_{s\in S}[D(s|H, X)] \nonumber + \frac{1}{|S|+|M|}\left(\sum_{j=1}^{|S|}[D(G(z_j)|H, X)] + \sum_{m\in M}[D(m|H, X)]\right)$. $|M|$ is a hyperparameter.} 
By keeping hard negative examples, we aim to prevent the discriminator from losing its generalization ability even if the generator accidentally samples only easy negative examples.

    \section{Experiments}
    \label{sec:exp}
    In this section, we review our experiments whose results support the effectiveness of AHP.
\begingroup
\setlength{\tabcolsep}{2.2pt} 
\renewcommand{\arraystretch}{1.2} 

\begin{table*}[ht]
    \vspace{-2mm}
    \centering
    \caption{\label{tab:performance} Hyperedge prediction performance and standard deviation. 
    In terms of AUROC averaged over four test sets (i.e., SNS, MNS, CNS, and S+M+C),
    AHP is up to \textbf{28.2\% better} than the best existing method and up to \textbf{5.5\% better} than its best variants.
    }
    
    \begin{subtable}[ht]{0.998 \textwidth}
    \centering
          \scalebox{0.69}{
            \begin{tabular}{c|cccc|a|cccc|a|cccc|a|cccc|a}
                \toprule
                Dataset &   \multicolumn{10}{c|}{\textbf{Citeseer}} 	& 	\multicolumn{10}{c}{\textbf{Cora-A}} \\
                \hline
                	&   	\multicolumn{5}{c|}{AUROC} 	& 	\multicolumn{5}{c|}{AP}	&   	\multicolumn{5}{c|}{AUROC} 	& 	\multicolumn{5}{c}{AP} 	\\		
            	\hline		
            	\rowcolor{white}
Test set	&	SNS	&	MNS	&	CNS	&	S+M+C	&	AVG	&	SNS	&	MNS	&	CNS	&	S+M+C	&	AVG	& SNS	&	MNS	&	CNS	&	S+M+C	&	AVG	&	SNS	&	MNS	&	CNS	&	S+M+C	&	AVG \\
\hline																					
Expansion&	0.663&	0.781&	0.331&	0.588&	0.591 $\pm$ 0.011&	0.765&	0.817&	0.498&	0.630&	0.681 $\pm$ 0.001 &  	0.690&	0.842&	0.434&	0.658&	0.656 $\pm$ 0.011&	0.690&	0.876&	0.577&	0.672&	0.706 $\pm$ 0.020 \\	
NHP&	\best{0.991}&	0.701&	0.510&	0.817&	0.751 $\pm$ 0.009&	\best{0.990}&	0.731&	0.520&	0.768&	0.751 $\pm$ 0.011 &  	0.909&	0.672&	0.550&	0.773&	0.723 $\pm$ 0.015&	0.925&	0.720&	0.585&	0.766&	0.748 $\pm$ 0.019 \\
HyperSAGNN&	0.540&	0.410&	0.473&	0.478&	0.475 $\pm$ 0.019&	0.627&	0.455&	0.497&	0.507&	0.521 $\pm$ 0.015 &  	0.386&	0.591&	0.542&	0.505&	0.506 $\pm$ 0.019&	0.532&	0.643&	0.545&	0.563&	0.571 $\pm$ 0.009 \\	
\hline											
AHP-S&	0.865&	0.844&	0.629&	0.785&	0.781 $\pm$ 0.030&	0.888&	0.838&	0.644&	0.778&	0.787 $\pm$ 0.027 &  	0.946&	0.917&	0.753&	0.870&	0.872 $\pm$ 0.011&	0.951&	0.896&	0.770&	0.857&	0.869 $\pm$ 0.016\\	
AHP-M&	0.825&	0.803&	0.613&	0.749&	0.748 $\pm$ 0.029&	0.837&	0.797&	0.639&	0.749&	0.756 $\pm$ 0.026 &  	0.940&	0.916&	0.758&	0.878&	0.873 $\pm$ 0.012&	0.945&	0.894&	0.785&	0.873&	0.874 $\pm$ 0.011\\		
AHP-C&	0.858&	0.844&	0.633&	0.774&	0.777 $\pm$ 0.019&	0.873&	0.839&	\best{0.661}&	0.773&	0.786 $\pm$ 0.013 &  	0.946&	0.918&	0.762&	0.880&	0.876 $\pm$ 0.014&	0.949&	0.896&	0.792&	0.875&	0.878 $\pm$ 0.014\\		
AHP-S+M+C&	0.861&	0.844&	0.629&	0.764&	0.774 $\pm$ 0.014&	0.873&	0.835&	0.653&	0.753&	0.779 $\pm$ 0.015 &  	0.945&	0.917&	0.762&	0.873&	0.874 $\pm$ 0.016&	0.950&	0.895&	\best{0.796}&	0.870&	0.878 $\pm$ 0.016\\	
\hline											
\textbf{AHP (Proposed)}&	0.943&	\best{0.881}&	\best{0.651}&	\best{0.820}&	\best{0.824 $\pm$ 0.020}&	0.952&	\best{0.870}&	0.660&	\best{0.795}&	\best{0.819 $\pm$ 0.022} &  	\best{0.958}&	\best{0.924}&	\best{0.782}&	\best{0.887}&	\best{0.888 $\pm$ 0.014}&	\best{0.957}&	\best{0.898}&	\best{0.796}&	\best{0.878}&	\best{0.882 $\pm$ 0.014}\\	
                \bottomrule
            \end{tabular}}
    \end{subtable}
    
    \begin{subtable}[ht]{0.998\textwidth}
    \centering
          \scalebox{0.69}{
            \begin{tabular}{c|cccc|a|cccc|a|cccc|a|cccc|a}
                \toprule
                Dataset &   \multicolumn{10}{c|}{\textbf{Cora}} 	& 	\multicolumn{10}{c}{\textbf{Pubmed}} \\
                \hline
                &	   	\multicolumn{5}{c|}{AUROC} 	& 	\multicolumn{5}{c|}{AP} &   	\multicolumn{5}{c|}{AUROC} 	& 	\multicolumn{5}{c}{AP} 	\\	
            	\hline			
            	\rowcolor{white}																		
Test set	&	SNS	&	MNS	&	CNS	&	S+M+C	&	AVG	&	SNS	&	MNS	&	CNS	&	S+M+C	&	AVG		&	SNS	&	MNS	&	CNS	&	S+M+C	&	AVG	&	SNS	&	MNS	&	CNS	&	S+M+C	&	AVG	\\
\hline																					
Expansion&	0.470&	0.707&	0.256&	0.476&	0.477 $\pm$ 0.009&	0.637&	0.764&	0.454&	0.563&	0.607 $\pm$ 0.009&	 	0.520&	0.730&	0.241&	0.497&	0.497 $\pm$ 0.015&	0.675&	0.755&	0.44&	0.565&	0.612 $\pm$ 0.010\\
NHP&	0.943&	0.641&	0.472&	0.774&	0.703 $\pm$ 0.015&	0.949&	0.678&	0.509&	\best{0.744}&	0.718 $\pm$ 0.020 &  	\best{0.973}&	0.694&	0.524&	0.745&	0.733 $\pm$ 0.004&	\best{0.973}&	0.656&	0.513&	0.678&	0.707 $\pm$ 0.004\\	
HyperSAGNN&	0.617&	0.527&	0.494&	0.540&	0.545 $\pm$ 0.021&	0.687&	0.574&	0.508&	0.566&	0.584 $\pm$ 0.019&  	0.525&	0.686&	0.546&	0.580&	0.584 $\pm$ 0.066&	0.534&	0.680&	0.529&	0.561&	0.576 $\pm$ 0.050\\	
\hline											
AHP-S&	0.935&	0.835&	0.565&	0.776&	0.777 $\pm$ 0.016&	0.939&	0.828&	\best{0.554}&	0.734&	0.764 $\pm$ 0.025&	0.904&	0.837&	0.535&	0.759&	0.759 $\pm$ 0.005&	0.909&	0.819&	0.524&	0.707&	0.740 $\pm$ 0.012\\	
AHP-M&	0.919&	0.836&	0.573&	0.782&	0.777 $\pm$ 0.012&	0.922&	0.823&	0.550&	0.737&	0.758 $\pm$ 0.018&	 	0.839&	\best{0.861}&	0.553&	0.751&	0.751 $\pm$ 0.010&	0.837&	\best{0.859}&	0.544&	0.712&	0.738 $\pm$ 0.017\\
AHP-C&	0.902&	0.824&	\best{0.575}&	0.765&	0.767 $\pm$ 0.024&	0.906&	0.813&	0.553&	0.726&	0.749 $\pm$ 0.040&	 	0.786&	0.839&	\best{0.555}&	0.726&	0.727 $\pm$ 0.007&	0.800&	0.842&	\best{0.553}&	0.702&	0.724 $\pm$ 0.011\\
AHP-S+M+C&	0.917&	0.830&	0.570&	0.769&	0.771 $\pm$ 0.018&	0.921&	0.820&	0.553&	0.719&	0.753 $\pm$ 0.026&	 	0.869&	0.855&	0.554&	0.762&	0.760 $\pm$ 0.010&	0.865&	0.856&	0.547&	\best{0.725}&	0.748 $\pm$ 0.008\\
\hline											
\textbf{AHP (Proposed)}&	\best{0.964}&	\best{0.860}&	0.572&	\best{0.799}&	\best{0.799 $\pm$ 0.019}&	\best{0.961}&	\best{0.837}&	0.552&	0.740&	\best{0.772 $\pm$ 0.035}&	 	0.917&	0.840&	0.533&	\best{0.763}&	\best{0.763 $\pm$ 0.009}&	0.918&	0.834&	0.526&	0.717&	\best{0.749 $\pm$ 0.007}\\

                \bottomrule
            \end{tabular}}
    \end{subtable}
    
    \begin{subtable}[ht]{0.998\textwidth}
    \centering
          \scalebox{0.69}{
            \begin{tabular}{c|cccc|a|cccc|a|cccc|a|cccc|a}
                \toprule
                Dataset &   \multicolumn{10}{c|}{\textbf{DBLP-A}} 	& 	\multicolumn{10}{c}{\textbf{DBLP}} \\
                \hline
                	&   	\multicolumn{5}{c|}{AUROC} 	& 	\multicolumn{5}{c|}{AP} 	&   	\multicolumn{5}{c|}{AUROC} 	& 	\multicolumn{5}{c}{AP} 	\\		
            	\hline			
            	\rowcolor{white}																		
                Test set	&	SNS	&	MNS	&	CNS	&	S+M+C	&	AVG	&	SNS	&	MNS	&	CNS	&	S+M+C	&	AVG	 	&	SNS	&	MNS	&	CNS	&	S+M+C	&	AVG	&	SNS	&	MNS	&	CNS	&	S+M+C	&	AVG	\\
                \hline
Expansion&	0.634&	0.826&	0.350&	0.603&	0.603 $\pm$ 0.006&	0.730&	0.852&	0.512&	0.641&	0.687 $\pm$ 0.004&  	0.645&	0.801&	0.366&	0.607&	0.607 $\pm$ 0.005&	0.751&	\best{0.856}&	0.518&	0.655&	0.698 $\pm$ 0.004\\
NHP&	\best{0.966}&	0.623&	0.555&	0.721&	0.716 $\pm$ 0.005&	\best{0.965}&	0.604&	0.534&	0.663&	0.693 $\pm$ 0.007&  	0.663&	0.540&	0.503&	0.572&	0.569 $\pm$ 0.003&	0.608&	0.523&	0.501&	0.542&	0.544 $\pm$ 0.002\\
HyperSAGNN&	0.548&	0.791&	0.563&	0.636&	0.634 $\pm$ 0.007&	0.686&	0.805&	0.552&	0.655&	0.675 $\pm$ 0.004&  	0.448&	0.574&	0.572&	0.530&	0.531 $\pm$ 0.018&	0.562&	0.602&	\best{0.586}&	0.577&	0.582 $\pm$ 0.016\\
\hline	
AHP-S	& 0.902	&	0.894	&	0.622	&	0.835	&	0.813	$\pm	0.011$	&	0.917	&	0.897	&	0.649	&	0.836	&	0.825	$\pm	0.006$	&  	0.944&	0.815&	0.557&	0.774&	0.773 $\pm$ 0.003&	0.944&	0.811&	0.546&	0.728&	0.757 $\pm$ 0.005\\
AHP-M	& 0.895	&	0.900	&	0.626	&	0.831	&	0.813	$\pm	0.012$&	0.910	&	0.904	&	0.656	&	0.833	&	0.826	$\pm	0.010$	& 	0.941&	\best{0.829}&	0.553&	0.773&	0.774 $\pm$ 0.001&	0.944&	0.829&	0.550&	0.733&	\best{0.764 $\pm$ 0.003}\\
AHP-C	& 0.891	&	0.893	&	0.615	&	0.833	&	0.812	$\pm	0.015	$	&	0.908	&	0.899	&	0.679	&	0.829	&	0.829	$\pm	0.011$	&  	0.876&	0.767&	0.566&	0.736&	0.736 $\pm$ 0.002&	0.875&	0.751&	0.563&	0.705&	0.723 $\pm$ 0.002\\ 
AHP-{S+M+C}	& 0.894	&	0.895	&	0.634	&	0.832	&	0.814	$\pm 0.011	$	&	0.911	&	0.901	&	0.680	&	0.835	&	0.832   $\pm 0.008$	&  	0.922&	0.805&	\best{0.573}&	0.767&	0.767 $\pm$ 0.017&	0.926&	0.800&	0.576&	\best{0.735}&	0.759 $\pm$ 0.018\\
                \hline																					
                \textbf{AHP (Proposed)} & 0.916	&	\best{0.926}	&	\best{0.668}	&	\best{0.838}	&	\best{0.837 $\pm$ 0.004}	&	0.928	&	\best{0.928}	&	\best{0.707}	&	\best{0.836}	&	\best{0.850 $\pm$ 0.003}	&
                 	\best{0.946}&	0.820&	0.568&	\best{0.778}&	\best{0.778 $\pm$ 0.002}&	\best{0.947}&	0.815&	0.561&	\best{0.735}&	\best{0.764 $\pm$ 0.007}\\
                \bottomrule
            \end{tabular}}
    \end{subtable}
    \vspace{3mm}
\end{table*}
\endgroup

\smallsection{Experimental settings.} We performed all experiments on a server with 256GB of RAM and RTX 2080 Ti GPUs, each of which has 11GB of vRAM. 
We split the hyperedges in each dataset into training (60\%), validation (20\%), and test (20\%) sets.
Validation and test hyperedges are unobserved and thus not used for node embedding.
We also masked some (spec., $1/6$) of training hyperedges during node embedding to facilitate generalization.
We formed four validation and test sets by adding  negative examples from SNS, MNS, CNS, and a mixture of them.
We measured AUROC and AP on each test set at an epoch when AUROC averaged over four validation sets was maximized, and we reported means over five runs.

\smallsection{Datasets} 
We used one \textbf{collaboration} (\textsf{DBLP}), three \textbf{co-citation} 
(\textsf{Citeseer}, \textsf{Cora}, and \textsf{Pubmed}), and
two \textbf{authorship} (\textsf{DBLP-A} and \textsf{Cora-A})  datasets.
See Appendix for some basic statistics of them.
\begin{itemize}[leftmargin=*]
    \item{\textbf{Collaboration dataset:}} A collaboration dataset is modeled as a hypergraph where each node is a researcher and each hyperedge is the set of (co-)authors of a paper.
    As in \cite{hypergcn}, we constructed the \textsf{DBLP} dataset from $22,964$ papers (i.e., hyperedges) from $87$ venues.\footnote{They belong to the following six conference categories: algorithms, database, programming, data mining, intelligence, and vision.}
    The papers were (co-)authored by $15,639$ researchers (i.e., nodes) and available at the Aminer academic network.\footnote{https://lfs.aminer.cn/lab-datasets/citation/DBLP-citation-network-Oct-19.tar.gz}
    For the features of each node, we averaged the bag-of-word features from the abstracts of the papers (co-)authored by the node.
    \item{\textbf{Co-citation datasets:}} A co-citation dataset is modeled as a hypergraph where each node is a paper and each hyperedge is the set of papers cited by a paper. For the features of each node, we used the bag-of-word features from its abstract.
    We used three co-citation datasets: \textsf{Citeseer}, \textsf{Cora}, and \textsf{Pubmed}.\footnote{https://linqs.soe.ucsc.edu/data}
    \item{\textbf{Authorship datasets:}} An authorship data is modeled as a hypergraph where each node is a paper and each hyperedge is the set of papers (co-)authored by a researcher. For the features of each node, we used the bag-of-word features from its abstract. We used two authorship datasets,  \textsf{DBLP-A}\footnote{https://github.com/malllabiisc/HyperGCN} and \textsf{Cora-A}\footnote{https://people.cs.umass.edu/~mccallum/data.html}, which were also used in \cite{hypergcn, dong2020hnhn}. 
\end{itemize}

\smallsection{Implementation details.}
We implemented AHP in PyTorch and Deep Graph Library (DGL).\footnote{https://www.dgl.ai/}
We set the node embedding dimension to $800$ for the \textsf{Pubmed} dataset and $400$ for the other datasets, as in \cite{dong2020hnhn}.
We used a three-layer MLP with LeakyReLU as activation functions for the generator. 
We set the dimension of MLP to $[64, 256, 256$, the number of nodes$]$ for the \textsf{Citeseer}, \textsf{Cora}, and \textsf{Cora-A} datasets and to $[128, 1024, 1024$, the number of nodes$]$ for the other datasets. 
We used a three-layer MLP with ReLU as activations functions for the classifier, and we set its dimension to $[$the node embedding dimension, $128, 8, 1]$. 
The discriminator and the generator were trained alternatingly for the same number of epochs.
We also tuned (a) the learning rates of the discriminator and the generator, (b) the normalization factors of HNHN, and (c) the size of the memory bank. See Appendix for their search space.

\smallsection{Baseline.}
As baseline approaches, we used
Expansion \cite{yoon2020much}, NHP \cite{nhp}, and HyperSAGNN \cite{zhang2019hyper}, which are state-of-the-art methods for hyperedge prediction formulated as a classification problem.
For them, we used the same node embedding dimension of AHP and used equal numbers of positive and negative examples in each epoch.
We tuned their learning rates among $\{0.01, 0.001, 0.0001\}$, and we set the other hyperparameters and sampled negative examples as suggested in the original publications.\footnote{For Expansion, we used features from 2- and 3-projected graphs and sampled negative examples so that the nodes in each example form a star in the 2-projected graph.}
We additionally considered four variants of AHP (AHP-S, AHP-M, AHP-C, and AHP-S+M+C) where only the learnable generator of AHP is replaced by SNS, MNS, CNS, and a mixture of them, respectively.
In them, negative examples are re-sampled at each iteration, as in AHP.
See Appendix for their hyperparameter search space.

\smallsection{Comparison with state of the arts.}
As shown in
Table~\ref{tab:performance}, AHP outperformed all state-of-the-art competitors (i.e., Expansion, NHP, and HyperSAGNN) on 19 (out of 24) and 17  test sets in terms of AUROC and AP, respectively. 
Moreover, \textbf{AHP performed best in all the six datasets in terms of AUROC and AP averaged over four test sets (i.e., SNS, MNS, CNS, and S+M+C).}
Especially, in the \textsf{DBLP} dataset, AHP was \textbf{28.2\% better}, in terms of average AUROC, than Expansion, which was the best existing method.
The generalization ability of Expansion, NHP, and HyperSAGNN, which all rely on heuristic sampling schemes, was very limited,
as seen in the fact that their AUROC was lower than $0.5$ on multiple test sets.

\smallsection{Comparison with variants.}
As shown in Table~\ref{tab:performance}, among the variants of AHP, one equipped with the same sampling scheme used for each test set tended to perform best.
For example, in the \textsf{Pubmed} dataset, AHP-S, AHP-M, AHP-C, and AHP-S+M+C performed best in test sets constructed by SNS, MNS, CNS, and SNS+MNS+CNS, respectively.
The results indicate that their generalization ability heavily depends on heuristic sampling schemes. 
\textbf{AHP performed better by up to $5.5\%$ than all its variants in terms of AUROC and AP averaged over four test sets, in all the six datasets.}
Moreover, in terms of AUROC, AHP even outperformed the variants equipped with the same sampling scheme used for each test set on 20 (out of 24) test sets.
The results confirm that our adversarial training scheme significantly improves the generalization ability.

\begin{figure}[t]
    \vspace{-2mm}
    \centering
    \includegraphics[width=\linewidth]{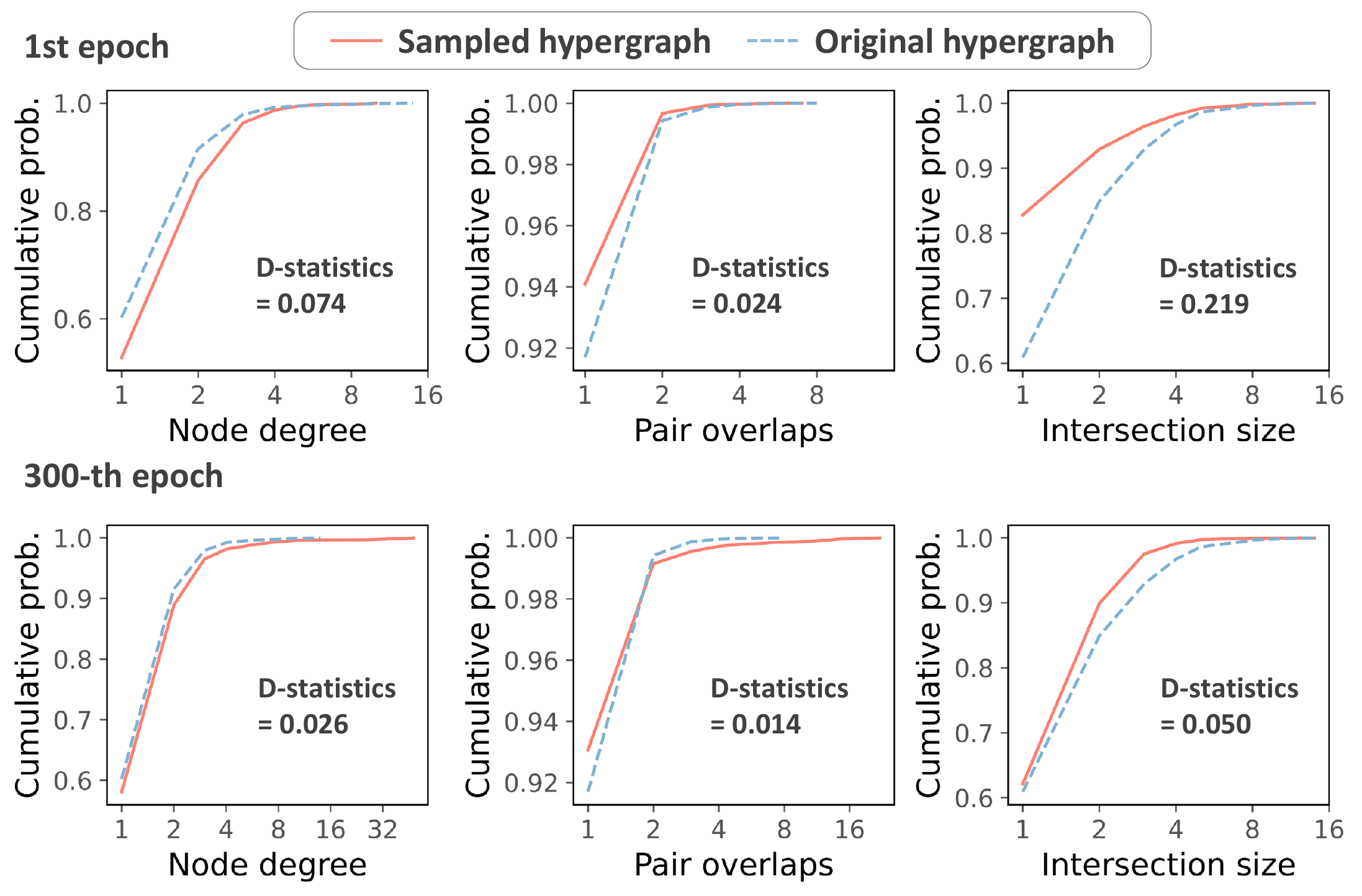} \\
    \vspace{-1mm}
    \caption{\label{fig:generator} Comparison of the hyperedges and the negative examples from the generator on the \textsf{Cora-A} dataset.
    Note that they became closer w.r.t. $3$ hypergraph measures  \cite{lee2021hyperedges}.
    }
\end{figure}

\smallsection{Analysis of the generator.}
In order to analyze how the the generator evolves during training, we compared the \textbf{original hypergraph} composed of the hyperedges in the training set and the \textbf{sampled hypergraph} where 
$1/6$ of the hyperedges in the original hypergraph were replaced by the same number of negative examples sampled by the generator.
Specifically, in them, we computed the distributions of (a) \textbf{node degree} (i.e., the number of hyperedges containing each node), (b) \textbf{pair overlaps} (i.e., the number of overlapping hyperedges on each pair of nodes), and (c) \textbf{intersection size} (i.e., the number of nodes in the intersection of each pair of hyperedges), which are suggested in \cite{lee2021hyperedges} as important hypergraph measures.
As shown in Fig.~\ref{fig:generator},  
the distributions in the original and sampled hypergraphs became closer to each other at the $300$-th epoch of training, compared to those at the first epoch.
We used Kolmogorov's D-statistic to measure the difference between distributions.
The results show that the generator  learned to sample negative examples closer  to positive examples.

\begin{figure}[t]
    \vspace{-2.5mm}
    \centering
    \includegraphics[width=\linewidth]{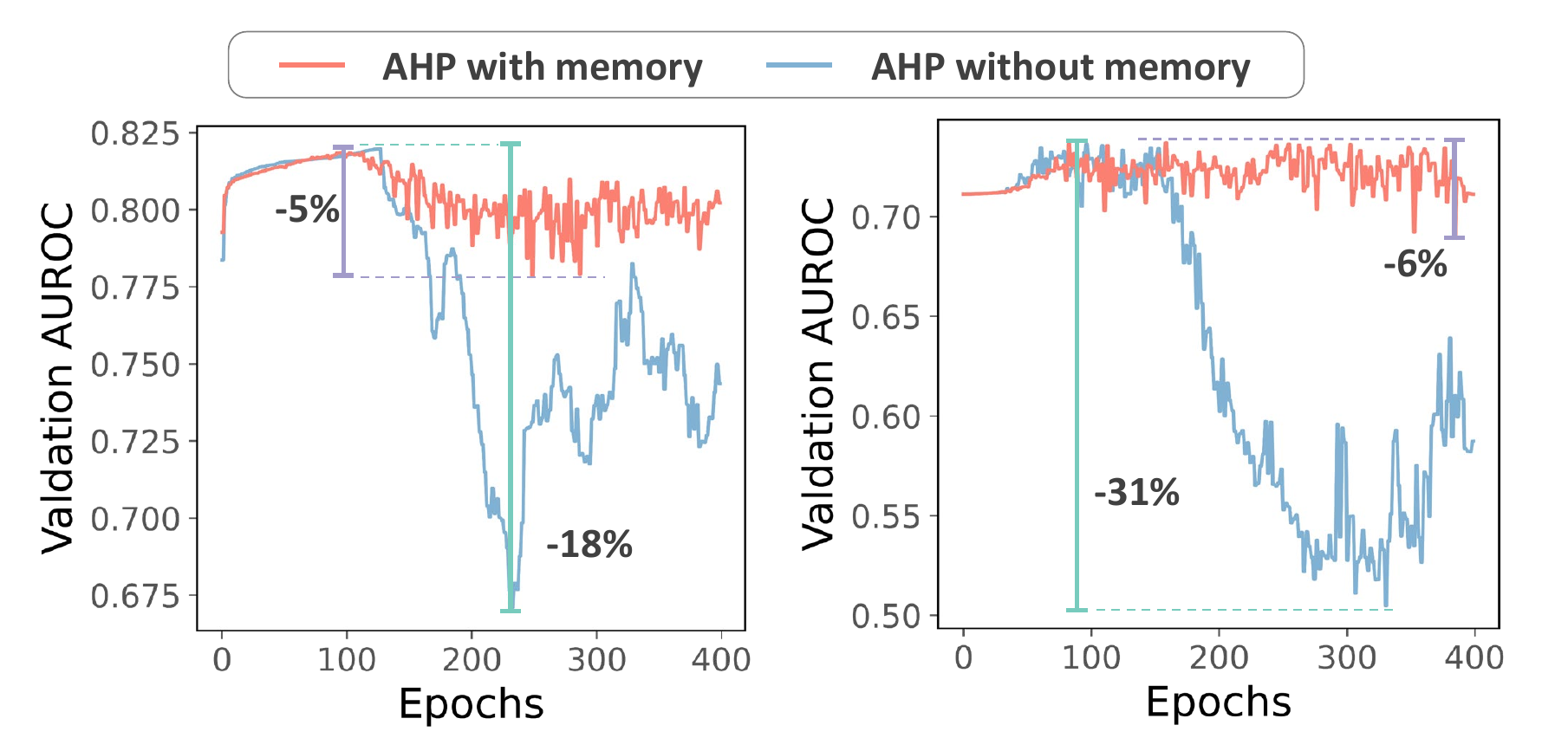} \\
    \vspace{-1mm}
    \caption{\label{fig:memory} The effects of the memory bank in the \textsf{Cora-A} (left) and 
    \textsf{DBLP} (right) datasets. The presented learning curves are averaged over five runs. Note that validation AUROC drops significantly without the memory bank.}
    
\end{figure}

\smallsection{Effects of the memory bank.}
In two datasets, using the memory bank helped stabilizing training, as shown in Fig.~\ref{fig:memory}.
However, its effect on final test AUROC was marginal in all datasets except for the \textsf{Pubmed} dataset, where final test AUROC improved by $1.1\%$.

    \section{Related works}
    \label{sec:related}
    \smallsection{Measure-based and projection-based approaches.}
Graph measures for edge prediction, including Katz's index \cite{Katz} and  common neighbors \cite{liben2007link}, have been extended to hypergraphs for hyperedge prediction \cite{zhang2018beyond}. 
Alternatively, Yoon et al. \cite{yoon2020much} projected a hypergraph into multiple pairwise graphs and fed the graph measures on them into a machine learning model for hyperedge prediction.
Zhang et al. \cite{zhang2018beyond} also projected a hypergraph into a pairwise graph via clique expansion and applied non-negative matrix factorization and least square matching for hyperedge prediction. This approach requires a feasible candidate set and thus inapplicable to Problem~\ref{def:HEpred}.

\smallsection{Deep learning-based approaches.}
Tu et al. \cite{tu2018structural} used a neural network to estimate hyperedge formation probabilities based on node embeddings. However, it is limited to hyperedges of a fixed size.
Zhang et al. \cite{zhang2019hyper} estimated hyperedge formation probabilities under the assumption that hyperedge-specific embeddings and general embeddings of nodes are similar within hyperedges.
The discriminator of AHP, including \textit{maxmin} pooling, is largely based on that used in \cite{nhp}, which applies a graph neural network \cite{kipf2016semi} (and a refining step) to clique expansion for node embedding, while our discriminator uses a hypergraph neural network \cite{dong2020hnhn} instead.

\smallsection{Relation to our work.} 
The learning-based approaches among the aforementioned ones require negative sampling, as discussed in Sect.~\ref{sec:intro}. To this end, they rely on heuristics, which can be replaced by our proposed adversarial training scheme.
In addition, for positive-unlabeled learning \cite{pu2017kiryo,na2020deep,wu2019long}, generative adversarial networks were used to generate fake positive and/or negative examples
\cite{zhou2021pure,hu2021predictive,hou2018generative}.
To the best of our knowledge, however, none of them was applied to hypergraphs or used to generate sets.

    \section{Conclusion}
    \label{sec:conclusion}
    In this paper, we proposed AHP, an adversarial training-based approach for hyperedge prediction.
Using six real-world hypergraphs, we revealed the limited generalization capability of state-of-the-art methods, which employ heuristics for negative sampling.
AHP replaces the heuristics with a learnable generator whose architecture is simple yet well-suited to hyperedge prediction. As a result, it yielded up to $28.2\%$ higher AUROC than the best existing method, generalizing better to negative examples of various natures.
Surprisingly, in most cases, AHP even outperformed its variants equipped with the same heuristics used to generate negative examples in test sets. For \textbf{reproducibility}, the code and datasets used in the paper are available at \url{https://github.com/HyunjinHwn/SIGIR22-AHP}.

\begin{table*}[t]
    \vspace{-2mm}
    \centering
    \caption{Details of the datasets.}
    \scalebox{0.8}{
        \begin{tabular}{c|c|c|c|c|c|c}
        \toprule
        	&	Citeseer	&	Cora	&	Cora-A	&	Pubmed  & DBLP-A & DBLP	\\
    	\midrule
    	Category &    Co-citation &    Co-citation &    Authorship &    Co-citation    & Authorship & Collaboration\\
        Number of nodes	&	1,457	&	1,434	&	2,388	&	3,840   & 39,283 & 15,639	\\
        Number of edges	&	1,078	&	1,579	&	1,072	&	7,962  & 16,483 & 22,964	\\
        Average size of hyperedges	&	3.2	&	3	&	4.3	&	4.3	  & 4.5 & 2.7\\
        Maximum size of hyperedges	&	26	&	5	&	43	&	171	  & 80 & 18\\
        Minimum size of hyperedges	&	2	&	2	&	2	&	2  & 2  & 2\\
        Dimension of node feature	&	3,703	&	1,433	&	1,433	&	500  & 4543 & 4543\\
        \bottomrule
        \end{tabular}}
    \label{tab:dataset}
\end{table*}

    \vspace{1mm}
    \smallsection{Acknowledgements:} This work was supported by National Research Foundation of Korea (NRF) grant funded by the Korea government (MSIT) (No. NRF-2020R1C1C1008296) and Institute of Information \& Communications Technology Planning \& Evaluation (IITP) grant funded by the Korea government (MSIT) (No. 2019-0-00075, Artificial Intelligence Graduate School Program (KAIST)).    
    
    \bibliographystyle{ACM-Reference-Format}
    \bibliography{ref}
    \balance

    \appendix
    \section*{Appendix}
    \begin{table}[t]
    \centering
    \caption{Hyperparameter search space}
    \label{tab:params}
    \begin{subtable}[ht]{0.48\textwidth}
    \centering
        \scalebox{0.8}{
            \begin{tabular}{cc}
        \toprule
        Hyperparameter & Selection pool \\
        \midrule
        Optimizer for the discriminator & Adam \\
        Optimizer for the generator & Adam \\
        Maximum epoch* & 400 \\
        Learning rate for the discriminator & 5e-03, 5e-04, 5e-05, 5e-06 \\
        Learning rate for the generator & 1e-04, 1e-05, 1e-06, 1e-07 \\
        Normalization factor $(\alpha, \beta)$** & (0,0), (1,1) \\
        Memory size & 0, 32, 128 \\
        \bottomrule
        \end{tabular}} 
    \caption{Search space for AHP}
    \label{tab:params(a)}
    \end{subtable}
    \begin{subtable}[ht]{0.48\textwidth}
    \centering
        \scalebox{0.8}{
            \begin{tabular}{cc}
        \toprule
        Hyperparameter & Selection pool \\
        \midrule
        Optimizer & Adam \\
        Learning rate & 5e-02, 5e-03, 5e-04, 5e-05, 5e-06 \\
        Normalization factor $(\alpha, \beta)$ & (0,0), (1,1) \\
        Maximum epoch* & 200 \\
        \bottomrule
        \multicolumn{2}{l}{\small{*Early stopped when the validation AUROC was maximized.}}\\
        \multicolumn{2}{l}{\small{**$\alpha$ and $\beta$ are degree normalization factors of hyperedges and nodes \cite{dong2020hnhn}.}}\\
        \end{tabular}} 
    \caption{Search space for the variants of AHP}
    \label{tab:params(b)}
    \end{subtable}
    \vspace{-3mm}
\end{table}

We provide in Table~\ref{tab:dataset} some basic statistics from the real-world hypergraphs used in the paper. We give in Table~\ref{tab:params} the hyperparameter search space of AHP and its variants.

\end{document}